\documentclass[letterpaper, 10 pt, journal, twoside]{IEEEtran}

\IEEEoverridecommandlockouts                              

\usepackage{graphicx}
\usepackage{caption}
\usepackage{amsmath,amssymb,enumerate}

\usepackage{amsthm}
\usepackage{mathtools}
\usepackage{breqn}
\usepackage{algorithm, algorithmicx, algpseudocode}

\usepackage{blindtext}
\usepackage{gensymb}
\usepackage{xparse}
\usepackage{lipsum}
\usepackage{mathrsfs}
\usepackage[mathscr]{euscript}
\usepackage{times}
\usepackage[noadjust]{cite} 
\usepackage{multicol}
\usepackage[caption=false,font=footnotesize]{subfig}
\usepackage{amsfonts}
\usepackage[T1]{fontenc}
\usepackage{textcomp}
\usepackage{balance}
\usepackage{amsfonts}
\usepackage{soul}
\usepackage{multirow}
\usepackage{booktabs}
\usepackage[usenames,dvipsnames,svgnames,table, xcdraw]{xcolor}

\usepackage[colorlinks=true,pdfpagemode=UseNone,citecolor=black,linkcolor=black,urlcolor=BrickRed]{hyperref}
\usepackage{comment}

\renewcommand{\thefigure}{\arabic{figure}}

\newcommand{\transpose}{\mathsf{T}}

\usepackage{pifont} 
\newcommand{\cmark}{\ding{51}}
\newcommand{\xmark}{\ding{55}}

\usepackage[capitalize]{cleveref}
\crefname{section}{Sec.}{Secs.}
\Crefname{section}{Section}{Sections}
\Crefname{table}{Table}{Tables}
\crefname{table}{Tab.}{Tabs.}

\begin{document}

\title{\LARGE \bf
SE3ET: SE(3)-Equivariant Transformer for Low-Overlap \\ Point Cloud Registration
}

\author{Chien Erh Lin, Minghan Zhu, and Maani Ghaffari
\thanks{Manuscript received: March 20, 2024; Revised June 6, 2024; Accepted July 2, 2024.}
\thanks{This paper was recommended for publication by Editor Sven Behnke upon evaluation of the Associate Editor and Reviewers’ comments.}
\thanks{This research was supported by NSF Award No. 2118818.}
\thanks{C.E. Lin, M. Zhu, and M. Ghaffari are with the University of Michigan, Ann Arbor, MI 48109, USA.
{\tt\small \{chienerh, minghanz,  maanigj\}@umich.edu}}%
\thanks{Digital Object Identifier (DOI): see top of this page.}
}

\markboth{IEEE Robotics and Automation Letters. Preprint Version. Accepted July, 2024}
{Lin \MakeLowercase{\textit{et al.}}: SE3ET: SE(3)-Equivariant Transformer for Low-Overlap Point Cloud Registration} 

\maketitle

\begin{abstract}

Partial point cloud registration is a challenging problem in robotics, especially when the robot undergoes a large transformation, causing a significant initial pose error and a low overlap between measurements. This work proposes exploiting equivariant learning from 3D point clouds to improve registration robustness. We propose SE3ET, an SE(3)-equivariant registration framework that employs equivariant point convolution and equivariant transformer designs to learn expressive and robust geometric features. We tested the proposed registration method on indoor and outdoor benchmarks where the point clouds are under arbitrary transformations and low overlapping ratios. We also provide generalization tests and run-time performance. 

\end{abstract}


\IEEEpeerreviewmaketitle

\section{Introduction}
\label{sec:introduction}
%
%
%
%
\IEEEPARstart{P}{oint} cloud registration has gained significant attention recently due to advancements in 3D sensor technology and computational resources. It seeks to determine the optimal transformation between two point clouds, addressing core challenges in computer vision, computer graphics, and robotics~\cite{clark2021nonparametric,zhang2021new}. These tasks include 3D localization, 3D reconstruction, pose estimation, and simultaneous localization and mapping (SLAM)~\cite{huang2021comprehensive}.

\begin{figure}
  \centering
  \includegraphics[width=0.95\columnwidth]{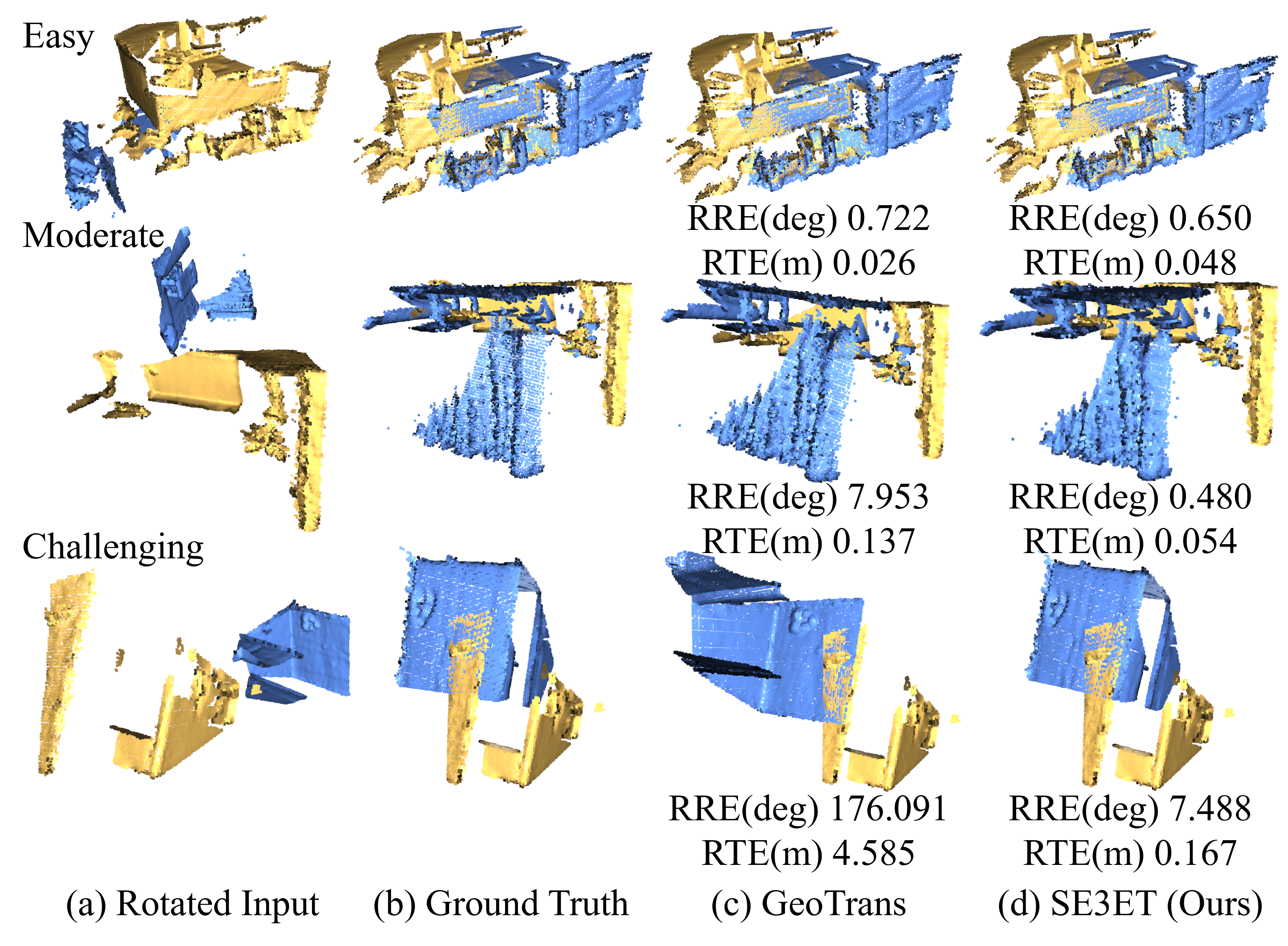}
  \caption{SE3ET can register two low-overlap point clouds with significant rotations and translations. This qualitative result is performed on rotated 3DLoMatch, where the first row is an easy example (28.72 \% overlapping ratio with multiple overlapping surfaces), the middle row is a moderate example (10.51 \% overlapping ratio with multiple overlapping surfaces), and the last row is a challenging example (26.75 \% overlapping ratio with only one overlapping surface).}
  \label{fig:qualitative_result}
\end{figure}

Partial-to-partial registration is widespread yet challenging in robotics applications. Many point cloud registration methods require sufficient overlap between two point clouds to find an accurate transformation~\cite{wang2022storm}. Recent works such as Predator~\cite{huang2021predator} and GeoTransformer~\cite{qin2022geometric} focus on low-overlap point cloud registration. However, these methods are not optimized for cases with significant initial pose errors, which are common in robotics. The current state-of-the-art works' limitations are shown in \cref{fig:qualitative_result}. 

Recent progress in equivariant networks for 3D point clouds~\cite{deng2021vectorneurons,chen2021epn,zhu2023e2pn,lin2023lie} enables neural networks to learn geometric features that preserve transformation information, providing a robust method to handle arbitrary transformations. We propose leveraging this advancement to improve point cloud registration solutions.

Few existing learning-based methods consider arbitrary transformations in the network architecture. For example, YOHO~\cite{wang2022yoho} applies equivariant feature learning using icosahedral-group convolution to learn rotation-equivariant point features. However, the framework can be further optimized for low-overlap registration. 

\begin{figure*}[t]
    \centering
    \includegraphics[width=0.95\textwidth]{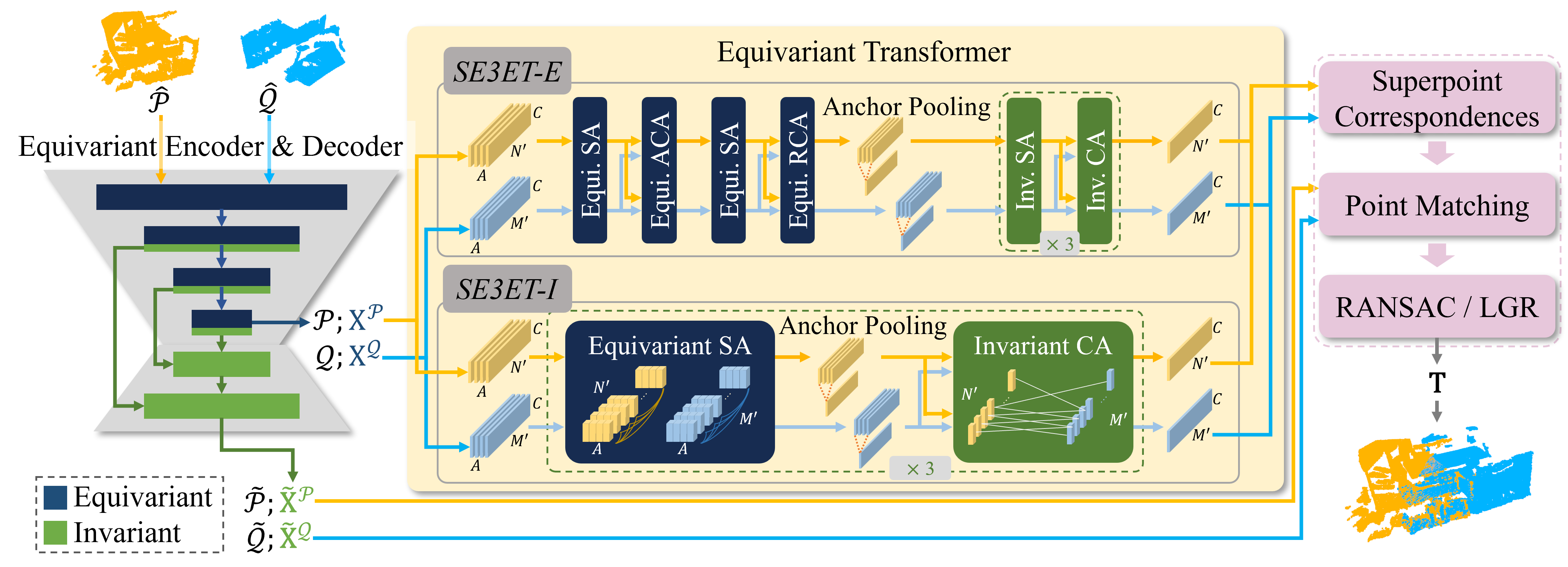}
    \caption{The proposed point cloud registration framework includes a $\mathrm{SE}(3)$-equivariant feature encoder and decoder and an equivariant transformer design for learning the point correspondences of superpoints. The dark blue blocks and arrows are equivariant, and the green blocks and arrows are invariant. We propose two transformers structures \textit{SE3ET-E} and \textit{SE3ET-I}. Here, SA stands for self-attention, and CA stands for cross-attention.}
    \label{fig:framework}
\end{figure*}

In this work, we propose SE3ET, a more robust and efficient $\mathrm{SE}(3)$-equivariant low-overlap point cloud registration framework. This framework (shown in \cref{fig:framework}) includes an equivariant feature learning encoder-decoder and an equivariant transformer module, fully leveraging the advantages of equivariant networks. Our main contributions are:
\begin{enumerate}[1.]
    \item Equivariant convolutions and transformers improve robustness to point clouds with low overlap and large pose changes.
    \item We propose four designs of equivariant transformers, each offering unique benefits.
    \item We leverage the octahedral rotation group to improve efficiency and performance. This approach can also accommodate voxel downsampling. 
    \item Open-source software is available at \\ \url{https://github.com/UMich-CURLY/SE3ET}.
\end{enumerate}

\section{Related Work}
\label{sec:related_work}
Traditional point cloud registration methods extract handcrafted local features to establish point-to-point correspondences. One of the pioneering works is the Iterative Closest Point (ICP) algorithm~\cite{besl1992icp}. However, these methods often converge to local minima and are ineffective for large-scale data sets due to their reliance on geometric properties.


Learning-based methods have gained attention for their impressive performance. Correspondence-based methods~\cite{yu2021cofinet, qin2022geometric, fu2021rgm, wang2019dcp, cao2021pcam, yew2020rpmnet} follow a two-step process: correspondence matching and finding the optimal transformation. First, they compute correspondences between source and target point clouds based on point distances or feature similarities. Then, they calculate the optimal rigid transformation (rotation and translation) to align the point pairs, using RANSAC~\cite{fischler1981ransac}, weighted SVD~\cite{besl1992icp, wang2019dcp}, or LGR~\cite{qin2022geometric}. Although widely used, these methods are prone to incorrect correspondences. We propose using equivariant features to improve accuracy in our correspondence-based framework.


Partial overlap between point clouds, especially with noisy data or occlusions, poses a significant challenge. Strategies to address this include using graph neural networks~\cite{huang2021predator, wang2022storm}, learning posterior probabilities of Gaussian mixture models (GMMs)~\cite{mei2023unsupervised}, characterizing LiDAR's non-repetitive scanning pattern~\cite{aijazi2024non}, and employing transformer mechanisms~\cite{vaswani2017attention} to identify overlapping points~\cite{wang2022storm, qin2022geometric, zhu2021ropnet, yew2022regtr}. Another approach utilizes multi-scale features to find overlapping patches~\cite{yu2021cofinet, qin2022geometric, huang2022robust}. PEAL~\cite{yu2023peal} integrates prior overlap estimation but requires an estimated transformation before registration. These methods often rely on data augmentation to handle substantial rotational transformations.


Some studies, such as G3DOA~\cite{zhao2022g3doa}, PPFNet~\cite{deng2018ppfnet}, and SpinNet~\cite{ao2021spinnet}, craft rotation-invariant features. 
RoITr~\cite{yu2023roitr} and BUFFER~\cite{ao2023buffer} further extend SpinNet~\cite{ao2021spinnet} module to perform rotation-invariant point matching. However, these crafted rotation-invariant features could be oversimplified, which may sacrifice expressiveness. On the other hand, equivariant representations preserve more expressive features, providing better cross-domain generalizability than invariant representations. 

Recent studies~\cite{deng2021vectorneurons, chen2021epn, pmlr-v164-zhu22b, zhu2023e2pn, lin2023lie} focusing on learning equivariant features from 3D point clouds offer valuable insights into the relevance of network architecture for point cloud registration across various transformation scenarios. Moreover,~\cite{fuchs2020se} and~\cite{chatzipantazis2022se} research into integrating equivariant learning within the transformer mechanism demonstrates its applicability to tasks such as classification and reconstruction. However, despite these advancements, there remains a scarcity of learning-based methods for point cloud registration that adequately address arbitrary transformation situations within the network architecture. YOHO~\cite{wang2022yoho}, a notable example, employs rotation-equivariant feature learning and has been extended to RoReg~\cite{wang2023roreg}. While these methods effectively leverage equivariant feature learning for point cloud registration, further optimization is needed to bolster processing robustness, particularly in scenarios with low overlap.

Building upon these varied algorithmic approaches, this paper proposes a novel framework that potentially resolves issues of low overlapping point clouds in registration procedures robust to arbitrary transformation. In this paper, the optimal performance of E2PN~\cite{zhu2023e2pn} in learning $\mathrm{SE}(3)$-equivariant features has been harnessed by incorporating it in our feature learning process. Improvements in feature capabilities are achieved via the transformer mechanism's implicit learning of the overlapping points. Leveraging equivariant and invariant features enables a more robust registration for point clouds under arbitrary transformation possible.

\section{Preliminary}
We provide preliminaries on equivariance, invariance, and E2PN architecture of learning $\mathrm{SE}(3)$-equivariant features.

\subsection{Equivariance and Invariance}
\textit{Invariance} is where the output of a function remains unchanged under transformations of the input, expressed as $f_{inv}(g \circ x) = f_{inv}(x)$. This is useful for tasks like classification, where different input transformations represent the same object.

\textit{Equivariance} is a more generalized form where the function's output undergoes the same transformation as the input, expressed as $f_{equ}(g \circ x) = g \circ f_{equ}(x)$. Equivariant representations are more expressive than invariant representations, offering better cross-domain generalizability.

\subsection{E2PN}
\label{sec:pre_e2pn}
\begin{figure}[t]
    \centering
    \includegraphics[width=0.99\columnwidth]{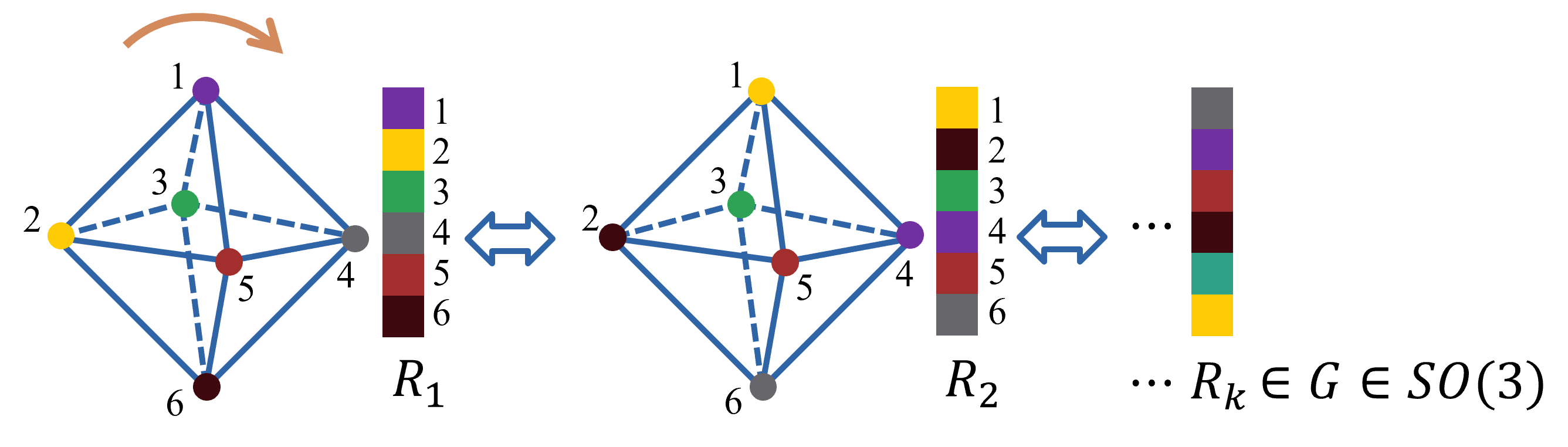}
    \caption{The geometric (octahedron shape) and algebraic (color bricks) illustration of using permutation to recover discretized rotation group. Each color represents the feature of one anchor (vertex), and the different order of the combination of features represents the discretized rotation defined in the network. If the octahedron is rotated 90 degrees clockwise, the order of the features in the anchor dimension changes accordingly. The discretization of the rotation groups is derived from the permutation of the discrete anchors. For $A = 6$, the rotation group contains 24 rotations.}
    \label{fig:permutation}
\end{figure}

In the SE3ET framework, E2PN~\cite{zhu2023e2pn} is used in the encoder and decoder modules to learn equivariant and invariant local features from 3D point clouds. 
E2PN extends Kernel Point Convolutions (KPConv)~\cite{thomas2019kpconv} by discretizing $\mathrm{SO}(3)$ into a polyhedral rotation group $G$, approximating $\mathrm{SE}(3)$ as $\mathbb{R}^3\times G$. Feature maps are defined on $\mathbb{R}^3\times V$, with $V$ representing polyhedral vertices, or \textit{anchors}. These maps are computed via quotient-space convolution. 
E2PN features have an additional anchor dimension. When a discretized rotation group transforms the point cloud, the features are permuted on the anchor dimension, as shown in \cref{fig:permutation}.
Conventional transformers are not optimal for E2PN features, motivating us to develop an equivariant transformer.

\section{Methodology}
\label{sec:method}

Point cloud registration computes a (rigid) transformation $\mathbf{T} = \{ (\mathbf{R}, \mathbf{t}) \mid \mathbf{R} \in \mathrm{SO}(3), \mathbf{t} \in \mathbb{R}^3 \}\in \mathrm{SE}(3)$ that aligns two given partially overlapped point clouds $\hat{\mathcal{P}} = \{ \mathbf{p}_i \in \mathbb{R}^3 \mid i=1, \dots,N \}$ and $\hat{\mathcal{Q}} =\{ \mathbf{q}_j \in \mathbb{R}^3 \mid j=1, \dots,M \}$. 

The proposed framework, shown in \cref{fig:framework}, consists of three main components. Initially, an $\mathrm{SE}(3)$-equivariant feature encoder (\cref{sec:encoder}) is employed to extract equivariant and invariant features from input point clouds at various resolutions. Subsequently, coarse-level features from two point clouds learn correlations with each other in an equivariant transformer (\cref{sec:transformer}), comprising interleaved self-attention and cross-attention modules for spatial and rotational feature enhancement within and between point clouds. Finally, both coarse and fine-level $\mathrm{SE}(3)$-invariant features are utilized for point matching in the registration module (\cref{sec:registration}).

\subsection{Equivariant Feature Encoder and Decoder}
\label{sec:encoder}
We adopt GeoTransformer's multi-stage encoder-decoder structure~\cite{qin2022geometric}, following a Feature Pyramid Network~\cite{lin2017feature} to extract multi-scale features from downsampled point clouds. Different from conventional approaches, we employ E2PN~\cite{zhu2023e2pn}, an $\mathrm{SE}(3)$-equivariant point convolutional network, for $\mathrm{SE}(3)$-equivariant convolution. Point clouds downscaled to the coarsest level (termed "superpoints") are denoted as $\mathcal{P} \in \mathbb{R}^{N' \times 3}$, $\mathcal{Q} \in \mathbb{R}^{M' \times 3}$, with their equivariant features $\mathbf{X}^{\mathcal{P}} \in \mathbb{R}^{N' \times A \times C}$, $\mathbf{X}^{\mathcal{Q}} \in \mathbb{R}^{M' \times A \times C}$, where $N', M'$ denote the downsampled points, $C$ represents the feature channels, and $A=|V|$ denotes the anchor size. These equivariant features undergo enhancement in an equivariant transformer. After each encoder stage, max-pooling along the anchor dimension yields $\mathrm{SE}(3)$-invariant features. These features are then passed into the decoder, facilitating the extraction of $\mathrm{SE}(3)$-invariant features $\widetilde{\mathbf{X}}^{\mathcal{P}}$ and $\widetilde{\mathbf{X}}^{\mathcal{Q}} \in \mathbb{R}^{N' \times C}$ for fine-level point clouds $\widetilde{\mathcal{P}}$ and $\widetilde{\mathcal{Q}}$, essential for fine point matching (\cref{sec:registration}).

Different from E2PN's icosahedral implementation ($A=12$), we implement the octahedral ($A=6$) finite rotation groups for more efficient computation. By design, we can express each rotation in the discretized rotation group $G$ using a permutation of the vertices $V$. A geometric illustration is in \cref{fig:permutation}. 

\subsection{Equivariant Transformer Design}
\label{sec:transformer}
The original transformer~\cite{vaswani2017attention} operates using dot products of two matrices. However, the equivariant features derived from the encoder are 3-dimensional tensors (comprising point, anchor, and feature dimensions), each with its own geometric representation. Flattening these tensors would disregard their intrinsic structure. We propose equivariant transformer designs that preserve the additional anchor dimension to address this. We propose three unique equivariant transformer designs, each crafted for different purposes: (1) learning attention weights on the point dimension while accounting for anchor features, (2) learning attention weights on the anchor dimension, and (3) learning attention weights for each discretized rotation group. Additionally, we propose an invariant transformer design that learns attention weights on the point dimension when the input features are invariant, ensuring stability to $\mathrm{SE}(3)$ transformations.

We propose equivariant self-attention and cross-attention modules to enhance features of superpoints by gathering information across various spatial locations and orientations. Self-attention modules facilitate feature interaction within a point cloud, while cross-attention modules enable feature communication between pairs of point clouds. Below, we detail each module design.

\subsubsection{Equivariant Self-Attention (ESA) Module}
\label{sec:self_attention}
We introduce an equivariant self-attention module, extending self-attention methods in~\cite{qin2022geometric} while ensuring equivariance to $\mathrm{SE}(3)$ transformations. This module allows consistent behavior under rigid body transformations. 

We will use the lower subscript to denote the feature after a certain layer for simplicity of notation. For example, $\mathbf{x}_{\texttt{SA}}$ is the feature acquired from the self-attention module.

Since the self-attention module is conducted per point cloud, we focus on operations within point cloud $\mathcal{P}$, which similarly apply to $\mathcal{Q}$. Equivariant superpoint features $\mathbf{X}^{\mathcal{P}}$ serve as query, key, and value inputs. 
We follow~\cite{qin2022geometric} to use geometric information to provide geometric structure embedding $\mathbf{P}^{\mathcal{P}}\in \mathbb{R}^{N' \times N' \times C}$.

For a point indexed $i$ in $\mathcal{P}$ at anchor coordinate $r$, attention between such elements is computed using trainable weight matrices $\mathbf{W}^Q, \mathbf{W}^K,\mathbf{W}^V$ and $\mathbf{W}^P \in \mathbb{R}^{C \times C}$. Output features $\mathbf{x}_{\texttt{SA}, ir}^{\mathcal{P}}$ are obtained via weighted summation over points. A subsequent feed-forward layer, as in~\cite{vaswani2017attention}, refines learned features, resulting in $\mathbf{X}_{\texttt{SA}}^{\mathcal{P}} \in \mathbb{R}^{N' \times A \times C}$. 
\begin{equation}
    \label{eq:equi_sa_score}
        \mathbf{a}_{\texttt{SA}, ir,jr} = \frac{(\mathbf{x}_{ir}\mathbf{W}^Q)(\mathbf{x}_{jr}\mathbf{W}^K + \mathbf{p}_{i,j}\mathbf{W}^P)^\transpose}{\sqrt{C}},
\end{equation}
\begin{equation}
    \label{eq:equi_sa}
        \mathbf{x}_{\texttt{SA}, ir}^{\mathcal{P}} = \sum_{j=1}^{N'}\text{Softmax}_j(\mathbf{a}_{\texttt{SA}, ir,jr}) \mathbf{x}_{jr}\mathbf{W}^V,
\end{equation}
Compared with conventional self-attention among the point features, our equivariant self-attention module allows different attention values at different anchor coordinates for the same pair of points, similar to multi-head self-attention, but with the equivariant property. 

\subsubsection{Invariant Cross-Attention (ICA) Module}
The cross-attention mechanism integrates two separate inputs, typically observed in point cloud registration tasks with paired point clouds $\mathcal{P}$ and $\mathcal{Q}$. We focus on operations within $\mathcal{P}$; analogous operations apply to $\mathcal{Q}$. 

In the ICA module, attention is conducted on $\mathrm{SE}(3)$-invariant features derived from equivariant self-attention features $\mathbf{X}_{\texttt{SA}}^{\mathcal{P}}$ and $\mathbf{X}_{\texttt{SA}}^{\mathcal{Q}}$ for $\mathcal{P}$ and $\mathcal{Q}$, respectively. Pooling on the anchor dimension yields invariant features $\mathbf{X}_{\texttt{SA-inv}}^{\mathcal{P}} \in \mathbb{R}^{N' \times C}$ and $\mathbf{X}_{\texttt{SA-inv}}^{\mathcal{Q}} \in \mathbb{R}^{M' \times C}$.

In the ICA module, features $\mathbf{X}_{\texttt{SA-inv}}^{\mathcal{P}}$ serve as queries, and $\mathbf{X}_{\texttt{SA-inv}}^{\mathcal{Q}}$ as keys. The attention value between point $i$ in $\mathcal{P}$ and point $j$ in $\mathcal{Q}$ is computed using trainable weight matrices. 
\begin{equation}
    \label{eq:ci_score}
        \mathbf{a}_{\texttt{ICA}, i,j} = \frac{(\mathbf{x}_{\texttt{SA-inv}, i}^{\mathcal{P}}\mathbf{W}^Q)(\mathbf{x}_{\texttt{SA-inv}, j}^{\mathcal{Q}}\mathbf{W}^K)^\transpose}{\sqrt{C}}
\end{equation}
Depending on the desired output—equivariant or invariant—the values can be equivariant features $\mathbf{X}_{\texttt{SA}}^{\mathcal{Q}}$ or invariant features $\mathbf{X}_{\texttt{SA-inv}}^{\mathcal{Q}}$.  
The collection of equivariant output features is denoted as $\mathbf{X}_{\texttt{ICA}}^{\mathcal{P}} \in \mathbb{R}^{N' \times A \times C}$, while invariant output features are denoted as $\mathbf{X}_{\texttt{ICA-inv}}^{\mathcal{P}} \in \mathbb{R}^{N' \times C}$.
Take $\mathbf{X}_{\texttt{ICA}, i}^{\mathcal{P}}$ for point $i$ in $\mathcal{P}$ as an example,

\begin{equation}
    \label{eq:ci}
        \mathbf{x}_{\texttt{ICA}, i}^{\mathcal{P}} = \sum_{j=1}^{M'}\text{Softmax}_j(\mathbf{a}_{\texttt{ICA}, i,j}) \mathbf{x}_{\texttt{SA}, j}^{\mathcal{Q}}\mathbf{W}^V
\end{equation}

\begin{figure}[t]
    \centering
    \includegraphics[width=0.95\columnwidth]{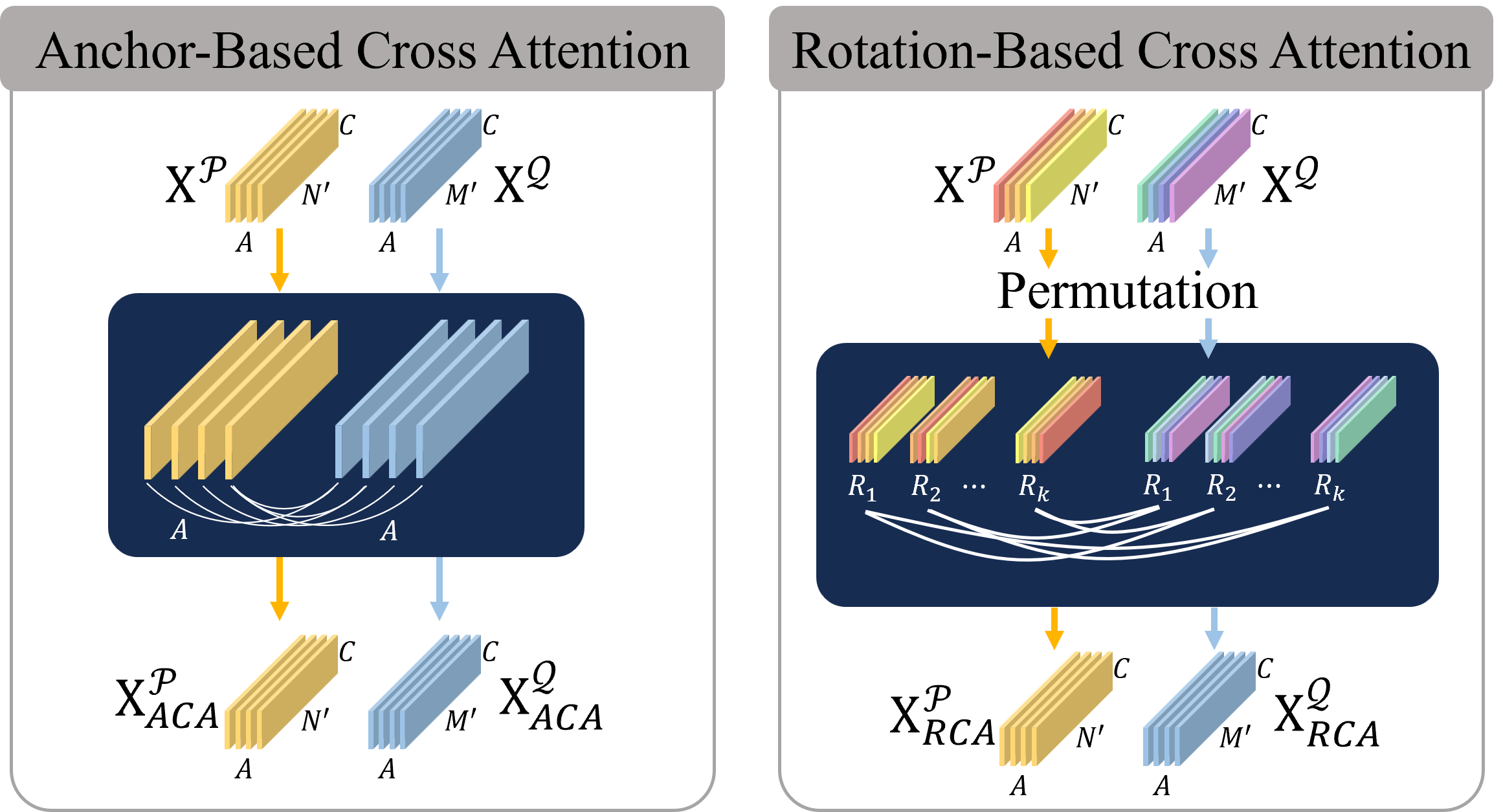}
    \caption{An illustration of the equivariant anchor-based cross-attention (ACA) and rotation-based cross-attention modules (RCA).}
    \label{fig:cross_attention}
\end{figure}

\subsubsection{Equivariant Anchor-Based Cross-Attention (ACA) Module}
In this module, we learn cross-attention scores for each anchor dimension. For point cloud $\mathcal{P}$, its equivariant feature from the self-attention module $\mathbf{X}_{\texttt{SA}}^{\mathcal{P}}$ serves as query and $\mathbf{X}_{\texttt{SA}}^{\mathcal{Q}}$ serves as key and value. Attention computation between two points in $\mathcal{P}$ and $\mathcal{Q}$ is performed via trainable weight matrices.

\begin{equation}
    \label{eq:equi_aca_raw}
        \mathbf{a}_{\texttt{ACA\_raw}, ir,js} = \frac{(\mathbf{x}_{\texttt{SA}, ir}^{\mathcal{P}}\mathbf{W}^Q)(\mathbf{x}_{\texttt{SA}, js}^{\mathcal{Q}}\mathbf{W}^K )^\transpose}{\sqrt{C}}
\end{equation}

Normalization is applied on both anchor and spatial dimensions to stabilize feature learning. Softplus ensures the non-negativity of raw attention, preventing false point-matching from distracting the anchor-wise attention between two point clouds. Global anchor-based attention, denoted as $\mathbf{a}_{\texttt{ACA\_anchor}}$, captures correlations between anchors across all points by average pooling on the point dimension. 
\begin{equation}
    \label{eq:equi_aca_anchor}
        \mathbf{a}_{\texttt{ACA\_anchor}, r,s} = \frac{1}{N'M'} \sum_{i=1}^{N'}\sum_{j=1}^{M'} \text{Softplus}(\mathbf{a}_{\texttt{ACA\_raw}, ir,js})
\end{equation}
We conduct normalization in both the anchor and spatial dimensions for stable feature learning. Anchor-wise, we calculate the normalized global anchor attention:
\begin{equation}
    \label{eq:equi_aca_norm}
        \mathbf{a}_{\texttt{ACA\_norm\_anchor}, r,s} = \frac{\mathbf{a}_{\texttt{ACA\_anchor}, r,s}}{\sum_{s=1}^{A}\mathbf{a}_{\texttt{ACA\_anchor}, r,s}}
\end{equation}
Spatial-wise, we apply softmax on the $j$ dimension to normalize the spatial dimension. 
\begin{equation}
    \label{eq:equi_aca_softmax}
        \mathbf{a}_{\texttt{ACA\_norm\_spatial}, ir,js} = \text{Softmax}_j(\mathbf{a}_{\texttt{ACA\_raw}, ir,js})
\end{equation}
We multiply \cref{eq:equi_aca_norm} and \cref{eq:equi_aca_softmax} to obtain the resulting attention for both the spatial and anchor dimensions. 
\begin{equation}
    \label{eq:equi_aca_all}
        \mathbf{a}_{\texttt{ACA}, ir,js} =  \mathbf{a}_{\texttt{ACA\_norm\_spatial}, ir,js} \mathbf{a}_{\texttt{ACA\_norm\_anchor},r,s}
\end{equation}

Output features for each point in $\mathcal{P}$ are obtained through weighted summation over points in $\mathcal{Q}$, followed by a feed-forward layer. 
\begin{equation}
    \label{eq:equi_aca}
        \mathbf{x}_{\texttt{ACA}, ir}^{\mathcal{P}} = \sum_{j=1}^{M'}\sum_{s=1}^{A} \mathbf{a}_{\texttt{ACA}, ir,js} \mathbf{x}_{\texttt{SA}, js}^{\mathcal{Q}}\mathbf{W}^V,
\end{equation}


ACA preserves equivariance, as the attention value depends solely on feature contents, regardless of anchor indices. This design ensures equivariant preservation, which is crucial for consistent behavior under rotations.

\subsubsection{Equivariant Rotation-Based Cross-Attention (RCA) Module}
In this version of cross-attention, we learn the cross-attention scores for each discretized rotation in the rotation group. 

First, we use the permutation layer from E2PN~\cite{zhu2023e2pn}, which is mentioned in \cref{sec:pre_e2pn} and \cref{sec:encoder}, to reconstruct feature maps defined on the discretized rotation group $G$ from the features defined on anchors $V$. We denote the permuted feature corresponding to the rotation $g\in G$ as: 
\begin{equation}\label{eq:permute}
    \mathbf{x}_{\texttt{Permute},jg}^{\mathcal{Q}}=\text{Permute}_g(\{\mathbf{x}_{\texttt{SA},js}^{\mathcal{Q}}\}_{s=1,...,A})
\end{equation}
After obtaining the feature corresponding to the rotation groups, the raw attention between two input features can be computed as:
\begin{equation}
    \label{eq:equi_rca_raw}
        \mathbf{a}_{\texttt{RCA\_raw}, i,jg} = \frac{(\mathbf{x}_{\texttt{SA}, i}^{\mathcal{P}}\mathbf{W}^Q)(\mathbf{x}_{\texttt{Permute},jg}^{\mathcal{Q}} \mathbf{W}^K )^\transpose}{\sqrt{C}}
\end{equation}
We carry out normalization in both the rotation and spatial dimensions for consistent feature learning. 
\begin{equation}\label{eq:equi_rca_rot}
        \mathbf{a}_{\texttt{RCA\_rot}, g} = \frac{1}{N'M'} \sum_{i=1}^{N'}\sum_{j=1}^{M'} \text{Softplus}(\mathbf{a}_{\texttt{RCA\_raw}, i,jg})
\end{equation}
\begin{equation}
    \label{eq:equi_rca_norm_rot}
    \mathbf{a}_{\texttt{RCA\_norm\_rot},g} = \frac{\mathbf{a}_{\texttt{RCA\_rot},g}}{\sum_{g=1}^{|P|}\mathbf{a}_{\texttt{RCA\_rot},g}}
\end{equation}
\begin{equation}
    \label{eq:equi_rca_softmax}
        \mathbf{a}_{\texttt{RCA\_norm\_spatial}, i,jg} = \text{Softmax}_j(\mathbf{a}_{\texttt{RCA\_raw}, i,jg})
\end{equation}
We multiply \cref{eq:equi_rca_norm_rot} and \cref{eq:equi_rca_softmax} to obtain attention for both the spatial and rotation dimensions. 
\begin{equation}
    \label{eq:equi_rca_all}
        \mathbf{a}_{\texttt{RCA}, i,jg} =  \mathbf{a}_{\texttt{RCA\_norm\_spatial}, i,jg} \ \mathbf{a}_{\texttt{RCA\_norm\_rot}, g}
\end{equation}
The output feature for the $i$'th point in $\mathcal{P}$, denoted as \mbox{$\mathbf{x}^{\mathcal{P}}_{\texttt{RCA}, i} \in \mathbb{R}^{A \times C}$}, can be written as:
\begin{equation}
    \label{eq:equi_rca_feat}
        \mathbf{x}^{\mathcal{P}}_{\texttt{RCA}, i} = \sum_{g=1}^{|P|}\sum_{j=1}^{M'} \mathbf{a}_{\texttt{RCA}, i,jg} (\mathbf{x}_{\texttt{Permute}, jg}^{\mathcal{Q}} \mathbf{W}^V)
\end{equation}
After passing through the feed-forward layer, we denote the collection of the output features as $\mathbf{X}_{\texttt{RCA}}^{\mathcal{P}} \in \mathbb{R}^{N' \times A \times C}$. 

\subsection{Equivariant Transformer Configurations}
To integrate the proposed transformer modules for point cloud registration, we present two configurations: the equivariant version, named \textit{SE3ET-E}, with all the proposed modules, and \textit{SE3ET-I}, which employs invariant cross-attention modules. Their structures are illustrated in \cref{fig:framework}.

SE3ET-E includes all the previously introduced modules, leveraging the full potential of equivariant features and optimizing performance where preserving geometric structure is crucial.

SE3ET-I combines equivariant self-attention with invariant cross-attention modules, reducing network complexity while maintaining $\mathrm{SE}(3)$-invariant attention learning to enhance registration robustness under arbitrary transformations. It is suitable for scenarios with large point clouds.

These configurations address different challenges: SE3ET-E for high geometric fidelity and SE3ET-I for robust transformation invariance and scalability.



\subsection{Registration and Estimate Transformation}
\label{sec:registration}
We adopt the methodology of GeoTransformer~\cite{qin2022geometric}, which consists of the following steps: superpoint-matching, fine point-matching, and a combined overlap-aware circle loss and point-matching loss for self-supervised learning. The output features from the transformer module are fed into the superpoint-matching module to identify patch correspondence. The fine-point matching module finds direct correspondence for a denser set of points. Afterward, transformation estimation is conducted either using RANSAC~\cite{fischler1981ransac} or a local-to-global (LGR) approach~\cite{qin2022geometric}. When utilizing RANSAC, a certain number of samples are chosen based on the high correspondence score derived from the fine-matching module. Lastly, the transformation between the two input point clouds is calculated via the Open3D library~\cite{zhou2018open3d}, considering the paired points.

\section{Experimental Results}
\label{sec:exp}
To show the generalizability of the proposed work, we conducted experiments on both indoor 3DMatch/3DLoMatch benchmark (\cref{sec:3dmatch_exp}) and outdoor KITTI point clouds (\cref{sec:kitti_exp}). We also tested the algorithm's generalizability (\cref{sec:generalize_exp}) and run-time performance (\cref{sec:runtime_exp}). 

\subsection{Indoor Benchmark: 3DMatch/3DLoMatch}
\label{sec:3dmatch_exp}
3DMatch~\cite{zeng20173dmatch} is a data set that provides noisy and partial RGB-D scanning data for indoor reconstruction. It consists of 62 scenes of point clouds, of which 46 are for training, 8 are for validation, and the other 8 scenes are for testing. 3DMatch contains point cloud pairs with over 30\% of overlapping rate. Predator~\cite{huang2021predator} extended it and created 3DLoMatch that only contains point cloud pairs with 10\% to 30\% overlap ratio for evaluating low overlapping point clouds. 

To show the proposed method's robustness under arbitrary pose changes, following YOHO~\cite{wang2022yoho}, we also report results where the point clouds in the testing set are transformed with \textit{random rotations} since the original data only contains limited pose changes. 

\subsubsection{Implementation Details}
\label{sec:3dmatch_implementation}
We use the octahedral rotation group with $A=6$. There are four stages in the encoder, and the equivariant features output from the encoder have feature size $C=1024$. SE3ET-E and SE3ET-I models are trained for 40 epochs with an initial learning rate of $10^{-4}$. 

\subsubsection{Evaluation Metrics}
\label{sec:3dmatch_metrics}
Following the literature~\cite{bai2020d3feat, huang2021predator, qin2022geometric}, we report \textit{Registration Recall (RR)}, as the registration metric. Given the estimated relative pose between a pair of point clouds, the RMSE of the locations of ground truth corresponding points between the aligned point clouds is calculated. RR is the percentage of point cloud pairs with RMSE smaller than a threshold (0.2 meters). This metric reflects the overall accuracy of the registration. 

For correspondence matching, we report \textit{Inlier Ratio (IR)}, defined as the proportion of estimated corresponding pairs of points of which the distance is smaller than a threshold (0.1 meters) under the ground truth transformation. This metric indicates the effectiveness of correspondence matching. A higher IR suggests a more precise match of correspondence between two point clouds. 

We also report \textit{Feature Matching Recall (FMR)}, calculating the percentage of point cloud pairs with an inlier ratio higher than a certain threshold (5\%). This metric gauges how well the method can recover the pose with high confidence. A higher FMR reflects a more effective correspondence-based registration process, assuming the outlier matchings can be filtered out during the transformation estimation process.

To better understand the method's performance, we evaluate using the following absolute metrics. \textit{Relative Translation Error (RTE)}: the Euclidean distance between the estimated translation and the ground truth translation vector. \textit{Relative Rotation Error (RRE)}: an isotropic error between the estimated rotation matrix $\hat{\mathbf{R}}$ and the ground truth rotation matrix $\mathbf{R}$, calculated as $\arccos(\frac{\mathrm{tr}(\hat{\mathbf{R}}^\transpose \mathbf{R} - 1)}{2})$. 

\subsubsection{Evaluation Results}
We compare the original and rotated 3DMatch/3DLoMatch result with state-of-the-art point cloud registration methods in \cref{tab:3dmatch_rebuttal_result}. We reproduce GeoTransformer's~\cite{qin2022geometric}, PEAL's~\cite{yu2023peal}, and CoFiNet's~\cite{yu2021cofinet} performance with their pre-trained weight. We report FCGF~\cite{choy2019fcgf}, D3Feat~\cite{bai2020d3feat}, Predator~\cite{huang2021predator}, SpinNet~\cite{ao2021spinnet}, and YOHO~\cite{wang2022yoho} result from~\cite{wang2022yoho}, and compare RoReg's~\cite{wang2023roreg}, BUFFER's~\cite{ao2023buffer}, and RoITr's~\cite{yu2023roitr} performance according to their paper. The registration is performed with 5000 sample points and 50k iterations for RANSAC except YOHO and RoReg, which only run for 1k iterations, as stated in their paper. 

From \cref{tab:3dmatch_rebuttal_result}, our method performs the best among all other methods in arbitrary rotation, regardless of the overlapping ratio, showing superior robustness to low overlap and arbitrary rotations. Although RoReg demonstrates slightly better performance on the original 3DMatch, it fails to sustain the advantage on 3DLoMatch. We exclude PEAL~\cite{yu2023peal} in the ranking because PEAL uses overlap prior, which requires estimated transformation as input by running GeoTransformer beforehand. We also exclude RoITr~\cite{yu2023roitr} in the ranking because it requires normal directions as input. When comparing these two methods that require extra information, despite our lower inlier ratio, which is an intermediate metric, we maintain competitive results on other metrics, including the final registration recall.

\begin{table}
    \caption{Evaluation on the original and rotated 3DMatch/3DLoMatch benchmark. Excluding the methods using extra information, the best result is shown in \textbf{bold}, and the second best is shown in the \underline{underline}. SE3ET-E2 and SE3ET-I2 are the same as our proposed structures, with only half of the feature size.}
    \label{tab:3dmatch_rebuttal_result}
    \centering
    \resizebox{\columnwidth}{!}{
    \begin{tabular}{c|c|c|cc|cc}
    \toprule
    Symmetry & Extra & & \multicolumn{2}{c|}{3DMatch} & \multicolumn{2}{c}{3DLoMatch} \\
    Category & Information & Methods & Original & Rotated & Original & Rotated  \\
    \hline
    \multicolumn{7}{c}{\textit{Registration Recall (\%) $\uparrow$}} \\    
    \hline
    \multirow{5}{*}{None} 
     & \xmark & FCGF~\cite{choy2019fcgf} & 85.1 & 84.8 & 40.1 & 40.8 \\ 
     & \xmark & D3Feat~\cite{bai2020d3feat} & 81.6 & 83.0 & 37.2 & 36.1 \\ 
     & \xmark & Predator~\cite{huang2021predator} & 89.0 & 88.4 & 59.8 & 57.7 \\ 
     & \xmark & GeoTrans~\cite{qin2022geometric} & 91.5 & 87.7 & 74.8 & 68.5 \\ 
     & \cmark & \textcolor{gray}{PEAL~\cite{yu2023peal}} & \textcolor{gray}{94.6} & \textcolor{gray}{-} & \textcolor{gray}{81.7} & \textcolor{gray}{-} \\ 
    \hline
    \multirow{2}{*}{Invariant} 
     & \xmark & SpinNet~\cite{ao2021spinnet} & 88.6 & 88.4 & 59.8 & 58.1 \\ 
     & \cmark & \textcolor{gray}{RoITr~\cite{yu2023roitr}} & \textcolor{gray}{91.9} & \textcolor{gray}{94.7} & \textcolor{gray}{74.8} & \textcolor{gray}{77.2} \\
    \hline
    \multirow{5}{*}{Equivariant}
     & \xmark & YOHO~\cite{wang2022yoho} & 90.8 & \underline{90.6} & 65.2 & 65.9 \\ 
     & \xmark & RoReg~\cite{wang2023roreg} & \textbf{92.9} & - & 70.3 & - \\ 
     & \xmark & BUFFER~\cite{ao2023buffer} & \textbf{92.9} & - & 71.8 & -\\ 
     & \xmark & \textit{SE3ET-E/\textcolor{gray}{SE3ET-E2} (Ours)} & 91.8/\textcolor{gray}{91.4} & 88.3/\textcolor{gray}{88.0} & \textbf{77.0}/\textcolor{gray}{74.3} & \textbf{72.0}/\textcolor{gray}{68.1} \\ 
     & \xmark & \textit{SE3ET-I/\textcolor{gray}{SE3ET-I2} (Ours)} & \underline{92.6}/\textcolor{gray}{91.8} & \textbf{90.9}/\textcolor{gray}{87.8} & \underline{75.9}/\textcolor{gray}{74.9} & \underline{70.8}/\textcolor{gray}{70.1} \\ 
    \hline
    \multicolumn{7}{c}{\textit{Inlier Ratio (\%) $\uparrow$}} \\
    \hline 
    \multirow{5}{*}{None}  & \xmark & FCGF~\cite{choy2019fcgf} & 56.8 & 56.2 & 21.4 & 21.6 \\ 
     & \xmark & D3Feat~\cite{bai2020d3feat} & 39.0 & 39.2 & 13.2 & 13.5 \\ 
     & \xmark & Predator~\cite{huang2021predator} & 58.0 & 58.2 & 26.7 & 26.2 \\ 
    
     & \xmark & GeoTrans~\cite{qin2022geometric} & 71.0 & 68.0 & 42.6 & 39.4 \\ 
     & \cmark & \textcolor{gray}{PEAL~\cite{yu2023peal}} & \textcolor{gray}{72.4} & \textcolor{gray}{-} & \textcolor{gray}{45.0} & \textcolor{gray}{-} \\ 
    \hline
    \multirow{2}{*}{Invariant} & \xmark & SpinNet~\cite{ao2021spinnet} & 47.5 & 47.2 & 20.5 & 20.1 \\ 
    
     & \cmark & \textcolor{gray}{RoITr~\cite{yu2023roitr}} & \textcolor{gray}{82.6} & \textcolor{gray}{82.3} & \textcolor{gray}{54.3} & \textcolor{gray}{53.2} \\
    \hline
    \multirow{5}{*}{Equivariant} & \xmark & YOHO~\cite{wang2022yoho} & 64.4 & 65.1 & 25.9 & 26.4 \\ 
     & \xmark & RoReg~\cite{wang2023roreg} & \textbf{81.6} & - & 39.6 & -\\ 
     & \xmark & \textit{SE3ET-E/\textcolor{gray}{SE3ET-E2} (Ours)} & \underline{72.5}/\textcolor{gray}{73.4} & \textbf{71.5}/\textcolor{gray}{71.0} & 44.3/\textbf{44.8} & \textbf{43.6}/\textcolor{gray}{42.4} \\ 
     & \xmark & \textit{SE3ET-I/\textcolor{gray}{SE3ET-I2} (Ours)} & 71.2/\textcolor{gray}{72.5} & \underline{69.2}/\textcolor{gray}{69.8} & 42.0/\underline{43.4} & 40.4/\underline{41.2} \\ 
    \hline
    \multicolumn{7}{c}{\textit{Feature Matching Recall (\%) $\uparrow$}} \\
    \hline 
    \multirow{5}{*}{None} 
     & \xmark & FCGF~\cite{choy2019fcgf} & 97.4 & 97.6 & 76.6 & 75.4 \\ 
     & \xmark & D3Feat~\cite{bai2020d3feat} & 95.6 & 95.5 & 67.3 & 67.6 \\ 
     & \xmark & Predator~\cite{huang2021predator} & 96.6 & 96.7 & 78.6 & 75.7 \\ 
     
     & \xmark & GeoTrans~\cite{qin2022geometric} & 98.4 & \underline{98.2} & 87.4 & 85.8 \\ 
     & \cmark & \textcolor{gray}{PEAL~\cite{yu2023peal}} & \textcolor{gray}{99.0} & \textcolor{gray}{-} & \textcolor{gray}{91.7} & \textcolor{gray}{-} \\ 
    \hline     
    \multirow{2}{*}{Invariant}
    & \xmark & SpinNet~\cite{ao2021spinnet} & 97.6 & 67.5 & 75.3 & 75.3 \\ 
     & \cmark & \textcolor{gray}{RoITr~\cite{yu2023roitr}} & \textcolor{gray}{98.0} & \textcolor{gray}{98.2} & \textcolor{gray}{89.6} & \textcolor{gray}{89.4} \\
    \hline
    \multirow{5}{*}{Equivariant}
     & \xmark & YOHO~\cite{wang2022yoho} & 98.2 & 98.1 & 79.4 & 79.2 \\ 
     & \xmark & RoReg~\cite{wang2023roreg} & 98.2 & - & 82.1 & - \\ 
     & \xmark & \textit{SE3ET-E  / \textcolor{gray}{SE3ET-E2} (Ours)} & \underline{98.5}/\textcolor{gray}{97.9} & \underline{98.2}/\textcolor{gray}{98.0} & \textbf{88.8}/\textcolor{gray}{86.8} & \textbf{88.0}/\textcolor{gray}{85.8} \\ 
     & \xmark & \textit{SE3ET-I / \textcolor{gray}{SE3ET-I2} (Ours)} & 98.3/\textbf{98.6} & \textbf{98.5}/\textcolor{gray}{97.9} & \underline{87.7}/\textcolor{gray}{88.4} & \underline{87.2}/\textcolor{gray}{87.8} \\ 
    \bottomrule
    \end{tabular}
    }
\end{table}

We further examine the result for using different numbers of samples for the original 3DMatch/3DLoMatch in \cref{tab:3dmatchsampleresult} and rotated 3DMatch/3DLoMatch \cref{tab:3dmatch_rotated_sample_result}. In \cref{tab:3dmatchsampleresult}, corresponding to the not-rotated testing set, our models yield the best performance in the 3DLoMatch testing set, showing the robust performance of our method in harsh low-overlapping conditions. In the 3DMatch testing set, where the overlap ratio is higher, we observe mixed results in the inlier ratio (IR). We hypothesize that local point cloud patches with rotational symmetry could obscure the rotation-invariant features. Because the not-rotated testing set only includes very limited relative pose changes between a point cloud pair, the non-equivariant baseline methods will not suffer from the pose ambiguity and can overfit the feature descriptor to the local pose difference between the local patches. Compared with other equivariant registration methods (YOHO~\cite{wang2022yoho} and RoReg~\cite{wang2023roreg}), our method performs better in most metrics. RoReg has slightly higher registration recalls in 3DMatch, but our method leads with a considerable margin in other metrics, especially in the 3DLoMatch benchmark. 

From \cref{tab:3dmatch_rotated_sample_result}, it could be observed that our method offers better overall performance across different sample sizes when the input point clouds are subject to arbitrary rotations. Our method outperforms both the non-equivariant baseline (GeoTransformer~\cite{qin2022geometric}) and the equivariant baseline (YOHO~\cite{wang2022yoho}). 
\cref{tab:3dMatch_registration_absolute} presents the 3DMatch and 3DLoMatch registration result in absolute metrics. Our method performs well under low overlap and arbitrary rotation.

\begin{table}[t]
    \addtolength{\tabcolsep}{-4pt}
    \caption{Evaluation results on 3DMatch and 3DLoMatch for different number of samples. RANSAC is run with 1k iterations for YOHO and RoReg and 50k iterations for other methods. * symbol means the method uses extra information, which is excluded from the ranking. The best result is in \textbf{bold}, and the second is in the \underline{underline}.}
    \label{tab:3dmatchsampleresult}
    \centering
    \resizebox{\columnwidth}{!}{
    \begin{tabular}{l|ccccc|ccccc}
    \toprule
      & \multicolumn{5}{c|}{3DMatch} & \multicolumn{5}{c}{3DLoMatch} \\
     \# Samples & 5000 & 2500 & 1000 & 500 & 250 & 5000 & 2500 & 1000 & 500 & 250 \\
    \hline
    \multicolumn{11}{c}{\textit{Registration Recall (\%) $\uparrow$}} \\
    \hline
    Predator~\cite{huang2021predator} & 89.8 & 89.9 & 89.8 & 87.3 & 87.1 & 60.2 & 61.1 & 62.7 & 61.3 & 57.4 \\ 
    CoFiNet~\cite{yu2021cofinet} & 90.2 & 90.7 & 90.6 & 90.2 & 89.5 & 67.2 & 66.2 & 68.5 & 66.1 & 61.3 \\ 
    YOHO~\cite{wang2022yoho} & 90.8 & 90.3 & 89.1 & 88.6 & 84.5 & 65.2 & 65.5 & 63.2 & 56.5 & 48.0 \\ 
    RoReg~\cite{wang2023roreg} & \textbf{92.9} & \textbf{93.2} & \textbf{93.2} & \textbf{94.2} & \textbf{92.0} & 70.3 & 70.2 & 68.3 & 67.6 & 64.9 \\ 
    GeoTrans~\cite{qin2022geometric} & 91.5 & 91.3 & 90.4 & 90.9 & 90.1 & 74.8 & 74.4 & 74.7 & \underline{73.7} & 73.2 \\
    BUFFER~\cite{ao2023buffer} & \textbf{92.9} & - & - & - & - & 71.8 & - & - & - & - \\ 
    \textcolor{gray}{PEAL*~\cite{yu2023peal}} & \textcolor{gray}{94.6} & \textcolor{gray}{93.7} & \textcolor{gray}{93.7} & \textcolor{gray}{93.9} & \textcolor{gray}{93.4} & \textcolor{gray}{81.7} & \textcolor{gray}{81.2} & \textcolor{gray}{80.8} & \textcolor{gray}{80.4} & \textcolor{gray}{80.1} \\
    \textcolor{gray}{RoITr*~\cite{yu2023roitr}} & \textcolor{gray}{91.9} & \textcolor{gray}{91.7} & \textcolor{gray}{91.8} & \textcolor{gray}{91.4} & \textcolor{gray}{91.0} & \textcolor{gray}{74.7} & \textcolor{gray}{74.8} & \textcolor{gray}{74.8} & \textcolor{gray}{74.2} & \textcolor{gray}{73.6} \\
    \textit{SE3ET-E (Ours)} & 91.8 & 91.9 & 91.1 & 91.5 & 90.4 & \textbf{77.0} & \textbf{77.0} & \textbf{76.2} & \textbf{75.6} & \textbf{75.0} \\ 
    \textit{SE3ET-I (Ours)} & \underline{92.6} & \underline{92.4} & \underline{92.2} & \underline{92.3} & \underline{91.5} & \underline{75.9} & \underline{74.7} & \underline{75.0} & 73.5 & \underline{74.0} \\ 
    \hline
    \multicolumn{11}{c}{\textit{Inlier Ratio (\%) $\uparrow$}} \\
    \hline
    Predator~\cite{huang2021predator} & 58.0 & 58.4 & 57.1 & 54.1 & 49.3 & 26.7 & 28.0 & 28.3 & 25.7 & 25.7 \\ 
    CoFiNet~\cite{yu2021cofinet} & 49.9 & 51.2 & 51.9 & 52.1 & 52.2 & 24.4 & 25.9 & 26.7 & 26.8 & 27.1 \\ 
    YOHO~\cite{wang2022yoho} & 64.4 & 60.7 & 55.7 & 46.4 & 41.2 & 25.9 & 23.3 & 22.6 & 18.2 & 15.0 \\ 
    RoReg~\cite{wang2023roreg} & \textbf{81.6}  & \textbf{80.7} & 75.1 & 74.6 & 76.0 & 39.6 & 39.9 & 33.6 & 32.0 & 34.5 \\ 
    GeoTrans~\cite{qin2022geometric} & 71.0 & 77.7 & \textbf{82.3} & \textbf{84.1} & \textbf{85.3} & \underline{42.6} & 47.9 & \underline{54.0} & \underline{56.6} & \textbf{58.4} \\ 
    \textcolor{gray}{PEAL*~\cite{yu2023peal}} & \textcolor{gray}{72.4} & \textcolor{gray}{79.1} & \textcolor{gray}{84.1} & \textcolor{gray}{86.1} & \textcolor{gray}{87.3} & \textcolor{gray}{45.0} & \textcolor{gray}{50.9} & \textcolor{gray}{57.4} & \textcolor{gray}{60.3} & \textcolor{gray}{62.2} \\
    \textcolor{gray}{RoITr*~\cite{yu2023roitr}} & \textcolor{gray}{82.6} & \textcolor{gray}{82.8} & \textcolor{gray}{83.0} & \textcolor{gray}{83.0} & \textcolor{gray}{83.0} & \textcolor{gray}{54.3} & \textcolor{gray}{54.6} & \textcolor{gray}{55.1} & \textcolor{gray}{55.2} & \textcolor{gray}{55.3} \\
    \textit{SE3ET-E (Ours)} & \underline{72.5} & \underline{77.8} & \underline{81.2} & \underline{82.5} & 83.2 & \textbf{44.3} & \textbf{50.2} & \textbf{54.8} & \textbf{56.7} & \underline{57.9} \\ 
    \textit{SE3ET-I (Ours)} & 71.2 & 77.1 & 80.9 & 82.4 & \underline{83.4} & 42.0 & \underline{48.1} & 53.1 & 55.2 & 56.7 \\ 
    \hline
    \multicolumn{11}{c}{\textit{Feature Matching Recall (\%) $\uparrow$}} \\
    \hline
    Predator~\cite{huang2021predator} & 96.6 & 96.3 & 96.4 & 96.6 & 96.6 & 78.1 & 79.3 & 79.4 & 79.2 & 77.8 \\ 
    CoFiNet~\cite{yu2021cofinet} & 98.1 & 98.2 & 98.2 & 98.2 & \underline{98.2} & 83.2 & 83.1 & 82.9 & 82.7 & 82.7 \\ 
    GeoTrans~\cite{qin2022geometric} & \underline{98.4} & \underline{98.4} & \underline{98.4} & 98.1 & 98.0 & 87.4 & 87.2 & 87.7 & \underline{87.8} & \underline{87.3} \\ 
    YOHO \cite{wang2022yoho} & 98.2 & 97.6 & 97.5 & 97.7 & 96.0 & 79.4 & 78.1 & 76.3 & 73.8 & 69.1 \\ 
    RoReg~\cite{wang2023roreg} & 98.2 & 98.0 & 98.3 & 98.2 & 97.8 & 82.1 & 82.3 & 81.2 & 80.7 & 80.5 \\ 
    \textcolor{gray}{PEAL*~\cite{yu2023peal}} & \textcolor{gray}{99.0} & \textcolor{gray}{99.0} & \textcolor{gray}{99.1} & \textcolor{gray}{99.1} & \textcolor{gray}{98.8} & \textcolor{gray}{91.7} & \textcolor{gray}{92.4} & \textcolor{gray}{92.5} & \textcolor{gray}{92.9} & \textcolor{gray}{92.7} \\
    \textcolor{gray}{RoITr*~\cite{yu2023roitr}} & \textcolor{gray}{98.0} & \textcolor{gray}{98.0} & \textcolor{gray}{97.9} & \textcolor{gray}{98.0} & \textcolor{gray}{97.9} & \textcolor{gray}{89.6} & \textcolor{gray}{89.6} & \textcolor{gray}{89.5} & \textcolor{gray}{89.4} & \textcolor{gray}{89.3} \\
    \textit{SE3ET-E (Ours)} & \textbf{98.5} & \textbf{98.7} & \textbf{98.6} & \textbf{98.4} & \textbf{98.3} & \textbf{88.8} & \textbf{88.6} & \textbf{88.6} & \textbf{88.5} & \textbf{88.0} \\ 
    \textit{SE3ET-I (Ours)} & 98.3 & 98.3 & 98.3 & \underline{98.3} & 98.1 & \underline{87.7} & \underline{87.7} & \underline{87.8} & 87.7 & 86.7 \\ 
    \bottomrule
    \end{tabular}
    }
\end{table}

\begin{table}[t]
    \addtolength{\tabcolsep}{-4pt}
    \caption{Evaluation results on rotated 3DMatch and 3DLoMatch for different number of samples. RANSAC is run with 1k iterations for YOHO, as stated in their paper, and 50k iterations for other methods.}
    \label{tab:3dmatch_rotated_sample_result}
    \centering
    \resizebox{\columnwidth}{!}{
    \begin{tabular}{l|ccccc|ccccc}
    \toprule
      & \multicolumn{5}{c|}{Rotated 3DMatch} & \multicolumn{5}{c}{Rotated 3DLoMatch} \\
     \# Samples & 5000 & 2500 & 1000 & 500 & 250 & 5000 & 2500 & 1000 & 500 & 250 \\
    \hline
    \multicolumn{11}{c}{\textit{Registration Recall (\%) $\uparrow$}} \\
    \hline
    YOHO~\cite{wang2022yoho} & 90.6 & - & - & - & - & 65.9 & - & - & - & - \\ 
    GeoTrans~\cite{qin2022geometric} & 87.7 & 88.2 & 88.4 & 87.5 & 87.9 & 68.5 & 67.8 & 68.0 & 66.7 & 66.5\\
    \textit{SE3ET-E (Ours)} & 89.3 & 89.4 & 88.2 & \textbf{89.3} & \textbf{89.1} & \textbf{72.0} & \textbf{71.1} & \textbf{71.3} & \textbf{70.6} & \textbf{69.7} \\ 
    \textit{SE3ET-I (Ours)} & \textbf{90.9} & \textbf{90.7} & \textbf{89.9} & 89.1 & 88.5 & 70.8 & 70.3 & 71.0 & 69.8 & 68.1 \\ 
    \hline
    \multicolumn{11}{c}{\textit{Inlier Ratio (\%) $\uparrow$}} \\
    \hline
    YOHO~\cite{wang2022yoho} & 65.1 & - & - & - & - & 26.4 & - & - & - & - \\ 
    GeoTrans~\cite{qin2022geometric} & 68.0 & 74.8 & 80.1 & 82.3 & 83.6 & 39.4 & 44.1 & 50.7 & 53.6 & 55.5\\
    \textit{SE3ET-E (Ours)} & \textbf{71.5} & \textbf{77.6} & \textbf{81.7} & \textbf{83.3} & \textbf{84.4} & \textbf{43.6} & \textbf{49.5} & \textbf{55.1} & \textbf{57.5} & \textbf{59.0} \\ 
    \textit{SE3ET-I (Ours)} & 69.2 & 75.7 & 80.1 & 81.9 & 83.1 & 40.4 & 46.2 & 51.8 & 54.4 & 56.2 \\ 
    \hline
    \multicolumn{11}{c}{\textit{Feature Matching Recall (\%) $\uparrow$}} \\
    \hline
    YOHO~\cite{wang2022yoho} & 98.1 & - & - & - & - & 79.2 & - & - & - & - \\ 
    GeoTrans~\cite{qin2022geometric} & 98.2 & 98.0 & 98.1 & 97.9 & 98.0 & 85.8 & 86.1 & 86.0 & 86.0 & 85.9\\
    \textit{SE3ET-E (Ours)} & 98.2 & \textbf{98.2} & 98.1 & 98.0 & 98.1 & \textbf{88.0} & \textbf{88.4} & \textbf{88.5} & \textbf{88.1} & \textbf{87.2} \\ 
    \textit{SE3ET-I (Ours)} & \textbf{98.5} & 98.0 & \textbf{98.2} & \textbf{98.4} & \textbf{98.2} & 87.2 & 87.1 & 86.6 & 86.5 & 87.0 \\ 
    \bottomrule
    \end{tabular}
    }
\end{table}

\begin{table}[t]
    \addtolength{\tabcolsep}{-4pt}
    \caption{Registration results on 3DMatch in absolute metrics Relative Translation Error (cm) and Relative Rotation Error (\textdegree).}  
    \scriptsize
    \label{tab:3dMatch_registration_absolute}
    \centering
    \resizebox{\columnwidth}{!}{
    \begin{tabular}{l|cc|cc|cc|cc}
    \toprule
        Methods & \multicolumn{2}{c|}{3DMatch} & \multicolumn{2}{c|}{Rotated 3DMatch} & \multicolumn{2}{c|}{3DLoMatch} & \multicolumn{2}{c}{Rotated 3DLoMatch} \\
        & RTE & RRE & RTE & RRE & RTE & RRE & RTE & RRE \\ 
        \hline
        GeoTrans~\cite{qin2022geometric} & 6.4 & \textbf{1.825} & 6.3 & 1.864 & 9.2 & 2.938 & 8.6 & 2.883 \\
        \textit{SE3ET-E (Ours)} & 6.4 & 1.893 & \textbf{6.1} & \textbf{1.792} & 9.0 & 2.973 & \textbf{8.3} & \textbf{2.762} \\ 
        \textit{SE3ET-I (Ours)} & 6.4 & 1.841 & 6.3 & 1.886 & \textbf{8.8} & \textbf{2.914} & 8.4 & 2.782 \\ 
    \bottomrule
    \end{tabular}}
\end{table}

\subsection{Outdoor Benchmark: KITTI Point Clouds}
\label{sec:kitti_exp}
In addition to indoor point clouds, we evaluate our framework on outdoor LiDAR point clouds from the KITTI odometry dataset~\cite{geiger2012kitti}. KITTI includes 11 sequences of LiDAR outdoor scans that are collected by Velodyne HDL-64E. Since this work focuses on addressing the low overlap point cloud registration challenge, we follow~\cite{qin2022geometric, huang2021predator, choy2019fully} to create point cloud pairs that are the first taken at least 10 meters apart within each sequence. Thus, this is not the complete KITTI odometry benchmark and the evaluation method is also different from standard KITTI odometry metrics. We use sequences 00-05 for training, 06-07 for validation, and 08-10 for testing.

\begin{table}[t]
    \caption{Evaluation result on KITTI point clouds.}
    \label{tab:kittiresult}
    \centering
    \resizebox{\columnwidth}{!}{
    \begin{tabular}{l|ccc}
    \toprule
     Method & RTE (cm) $\downarrow$ & RRE (\textdegree) $\downarrow$ & RR (\%) $\uparrow$ \\
    \hline
     FCGF~\cite{choy2019fcgf} & 9.5 & 0.30 & 96.6 \\ 
     D3Feat~\cite{bai2020d3feat} & 7.2 & 0.30 & \textbf{99.8} \\ 
     Predator~\cite{huang2021predator} & 6.8 & 0.27 & \textbf{99.8} \\ 
     CoFiNet~\cite{yu2021cofinet} & 8.2 & 0.41 & \textbf{99.8} \\ 
     SpinNet~\cite{ao2021spinnet} & 9.9 & 0.47 & 99.4 \\ 
     BUFFER~\cite{ao2023buffer} & 5.4 & 0.22 & -\\ 
     GeoTrans-RANSAC-50k~\cite{qin2022geometric} & 7.4 & 0.27 & \textbf{99.8} \\ 
     GeoTrans-LGR~\cite{qin2022geometric} & 6.8 & 0.24 & \textbf{99.8} \\ 
     \textit{SE3ET-I-RANSAC50k (Ours)} & 6.3 & 0.25 & 99.7 \\ 
     \textit{SE3ET-I-LGR (Ours)} & \textbf{4.7} & \textbf{0.21} & 99.7 \\ 
    \bottomrule
    \end{tabular}
    }
\end{table}



\begin{table}[t]
    \caption{Result for the generalization experiment. Evaluation results on KITTI point clouds with models trained on 3DMatch.}
    \footnotesize
    \label{tab:generalize_result}
    \centering
    \begin{tabular}{l|ccc}
    \hline
     Method & RTE (cm) $\downarrow$ & RRE (\textdegree) $\downarrow$ & RR (\%) $\uparrow$ \\
    \hline     
     GeoTrans & 42 & 1.26 & 70.6 \\     
     RoITr & 30 & 1.52 & 20.9 \\ 
     \textit{SE3ET-I2 (Ours)} & \textbf{17} & \textbf{0.71} & \textbf{75.9} \\ 
    \hline
    \end{tabular}
\end{table}

\begin{table}[th!]
    \caption{Runtime analysis of the proposed method.  
    We tested on an Intel i9-10900K CPU and Nvidia GeForce RTX 3090 GPU.}
    \label{tab:computation}
    \centering
    \resizebox{\columnwidth}{!}{
    \begin{tabular}{c|ccc|c}
    \hline
      & \multicolumn{3}{c|}{Number of Parameters} & Runtime \\
     Method & Backbone & Transformer & Total & (ms)\\
    \hline
     GeoTrans~\cite{qin2022geometric} & 6.01 M & 3.82 M & 9.83 M & 84\\
     RoITr~\cite{yu2023roitr} & 5.21 M & 4.81 M & 10.02 M & 232 \\ 
     \textit{SE3ET-E / \textcolor{gray}{SE3ET-E2} (Ours)} & 10.10 M / \textcolor{gray}{2.53 M} & 6.98 M / \textcolor{gray}{1.75 M} & 17.08 M/ \textcolor{gray}{4.29 M} & 210 / \textcolor{gray}{136}\\ 
     \textit{SE3ET-I \textcolor{gray}{SE3ET-I2} (Ours)} & 10.10 M / \textcolor{gray}{2.53 M} & 3.82 M / \textcolor{gray}{0.96 M} & 13.92 M / \textcolor{gray}{3.49 M} & 157 / \textcolor{gray}{105} \\ 
    \hline
    \end{tabular}}
\end{table}

For training on KITTI, we use the octahedral rotation group and set $A=6$. There are five stages in the encoder, and the equivariant features output from the encoder have feature size $C=2048$. In the transformer, we use the SE3ET-I design described in \cref{sec:transformer}, three equivariant self-attention layers, and three invariant cross-attention layers interleaved together. The model is trained for 160 epochs with an initial learning rate of $10^{-4}$.

Again, we evaluate the methods using RTE and RRE.
In addition, we evaluate using \textit{Registration Recall (RR)}: the percentage of point cloud pairs of which the RRE is smaller than 5 degrees and the RTE is smaller than 2 meters.

We compare with state-of-the-art point cloud registration methods in \cref{tab:kittiresult}. Among other methods, our approach delivers the smallest RTE and RRE. Although we didn't achieve the top performance in the RR metric, we are only 0.1 percentage points behind, with the RR nearing its saturation point at 100\%. It shows the superior performance of our method in outdoor driving scenes.

\subsection{Generalization Experiment}
\label{sec:generalize_exp}
In addition to the indoor and outdoor experiments, we aim to test how the proposed method generalizes to unseen data. Thus, we evaluate the pre-trained weights learned from the indoor RGB-D point cloud (3DMatch) in \cref{sec:3dmatch_exp} on outdoor LiDAR point clouds from KITTI. To address scaling issues between the two sensor measurements, we discard points beyond the 30-meter range and scale the KITTI point clouds by a factor of 0.1 to match the 3DMatch point cloud scale.

We evaluate the performance using RTE and RRE. The \textit{Registration Recall (RR)} is calculated as the percentage of scaled point cloud pairs of which the RRE is smaller than 5 degrees and the RTE is smaller than 0.2 meters (0.1 scale of the original threshold for the KITTI experiment).

We present the result in \cref{tab:generalize_result} 
 with the original KITTI point cloud scale to better compare with the result in \cref{tab:kittiresult}. Our equivariant features allow better generalizability in cross-dataset evaluation since equivariant representations preserve more expressive features and thus provide better cross-domain generalizability than non-equivariant and invariant representations. 


\subsection{Run-time Analysis}
\label{sec:runtime_exp}
In this section, We include details on network parameters and runtime for the proposed network. The result is shown in \cref{tab:computation}. Our proposed networks have faster runtime and fewer parameters than RoITr~\cite{yu2023roitr}. We can further increase efficiency by reducing the feature dimension by half in SE3ET-E2 and SE3ET-I2 (feature size $C=512$) while maintaining similar performance (performance shown in \cref{tab:3dmatch_rebuttal_result}). SE3ET-E2 and SE3ET-I2 use fewer parameters than GeoTrans~\cite{qin2022geometric} and still outperform it.

\section{Conclusion}
\label{sec:conclusion}

We have designed \textit{SE3ET}, a low-overlap point cloud registration framework that leverages $\mathrm{SE}(3)$-equivariant feature learning. Our approach enhances robustness to large transformations in low-overlap scenarios. We also propose an octahedral rotation group implementation and an equivariant transformer design to enable efficient training and improve performance. Experimental results on indoor and outdoor benchmarks demonstrate promising results across various metrics. Future work includes using the correlation between the anchor dimension of the equivariant features to obtain a coarse estimate of the rotation and reduce computation time for RANSAC. Moreover, combining this work within an equivariant place recognition framework~\cite{pmlr-v205-lin23a} for mobile robots is an attractive future direction. 

{
    \balance
    \small
    \bibliographystyle{IEEEtran}
    \bibliography{string-abrv, reference}

\begin{thebibliography}{10}
\providecommand{\url}[1]{#1}
\csname url@rmstyle\endcsname
\providecommand{\newblock}{\relax}
\providecommand{\bibinfo}[2]{#2}
\providecommand\BIBentrySTDinterwordspacing{\spaceskip=0pt\relax}
\providecommand\BIBentryALTinterwordstretchfactor{4}
\providecommand\BIBentryALTinterwordspacing{\spaceskip=\fontdimen2\font plus
\BIBentryALTinterwordstretchfactor\fontdimen3\font minus \fontdimen4\font\relax}
\providecommand\BIBforeignlanguage[2]{{%
\expandafter\ifx\csname l@#1\endcsname\relax
\typeout{** WARNING: IEEEtran.bst: No hyphenation pattern has been}%
\typeout{** loaded for the language `#1'. Using the pattern for}%
\typeout{** the default language instead.}%
\else
\language=\csname l@#1\endcsname
\fi
#2}}

\bibitem{clark2021nonparametric}
W.~Clark, M.~Ghaffari, and A.~Bloch, ``Nonparametric continuous sensor registration,'' \emph{J. Mach. Learning Res.}, vol.~22, no. 271, pp. 1--50, 2021.

\bibitem{zhang2021new}
R.~Zhang, T.-Y. Lin, C.~E. Lin, S.~A. Parkison, W.~Clark, J.~W. Grizzle, R.~M. Eustice, and M.~Ghaffari, ``A new framework for registration of semantic point clouds from stereo and {RGB-D} cameras,'' in \emph{Proc. {IEEE} Int. Conf. Robot. and Automation}.\hskip 1em plus 0.5em minus 0.4em\relax IEEE, 2021, pp. 12\,214--12\,221.

\bibitem{huang2021comprehensive}
X.~Huang, G.~Mei, J.~Zhang, and R.~Abbas, ``A comprehensive survey on point cloud registration,'' \emph{arXiv preprint arXiv:2103.02690}, 2021.

\bibitem{wang2022storm}
Y.~Wang, C.~Yan, Y.~Feng, S.~Du, Q.~Dai, and Y.~Gao, ``{STORM}: Structure-based overlap matching for partial point cloud registration,'' \emph{{IEEE} Trans. Pattern Anal. Mach. Intell.}, 2022.

\bibitem{huang2021predator}
S.~Huang, Z.~Gojcic, M.~Usvyatsov, A.~Wieser, and K.~Schindler, ``Predator: Registration of 3d point clouds with low overlap,'' in \emph{Proc. {IEEE} Conf. Comput. Vis. Pattern Recog.}, 2021, pp. 4267--4276.

\bibitem{qin2022geometric}
Z.~Qin, H.~Yu, C.~Wang, Y.~Guo, Y.~Peng, and K.~Xu, ``Geometric transformer for fast and robust point cloud registration,'' in \emph{Proc. {IEEE} Conf. Comput. Vis. Pattern Recog.}, 2022, pp. 11\,143--11\,152.

\bibitem{deng2021vectorneurons}
C.~Deng, O.~Litany, Y.~Duan, A.~Poulenard, A.~Tagliasacchi, and L.~J. Guibas, ``Vector neurons: A general framework for {SO(3)}-equivariant networks,'' in \emph{Proc. {IEEE} Int. Conf. Comput. Vis.}, 2021, pp. 12\,200--12\,209.

\bibitem{chen2021epn}
H.~Chen, S.~Liu, W.~Chen, H.~Li, and R.~Hill, ``Equivariant point network for 3d point cloud analysis,'' in \emph{Proc. {IEEE} Conf. Comput. Vis. Pattern Recog.}, 2021, pp. 14\,514--14\,523.

\bibitem{zhu2023e2pn}
M.~Zhu, M.~Ghaffari, W.~A. Clark, and H.~Peng, ``{E2PN}: Efficient {SE(3)}-equivariant point network,'' in \emph{Proc. {IEEE} Conf. Comput. Vis. Pattern Recog.}, 2023.

\bibitem{lin2023lie}
T.-Y. Lin, M.~Zhu, and M.~Ghaffari, ``{Lie Neurons}: Adjoint-equivariant neural networks for semisimple {Lie} algebras,'' in \emph{International Conference on Machine Learning}, 2024.

\bibitem{wang2022yoho}
H.~Wang, Y.~Liu, Z.~Dong, and W.~Wang, ``You only hypothesize once: Point cloud registration with rotation-equivariant descriptors,'' in \emph{Proceedings of the 30th ACM International Conference on Multimedia}, 2022, pp. 1630--1641.

\bibitem{besl1992icp}
P.~Besl and N.~McKay, ``A method for registration of 3-d shapes,'' \emph{{IEEE} Trans. Pattern Anal. Mach. Intell.}, vol.~14, no.~02, pp. 239--256, 1992.

\bibitem{yu2021cofinet}
H.~Yu, F.~Li, M.~Saleh, B.~Busam, and S.~Ilic, ``Cofinet: Reliable coarse-to-fine correspondences for robust pointcloud registration,'' \emph{Advances in Neural Information Processing Systems}, vol.~34, pp. 23\,872--23\,884, 2021.

\bibitem{fu2021rgm}
K.~Fu, S.~Liu, X.~Luo, and M.~Wang, ``Robust point cloud registration framework based on deep graph matching,'' in \emph{Proc. {IEEE} Conf. Comput. Vis. Pattern Recog.}, 2021, pp. 8893--8902.

\bibitem{wang2019dcp}
Y.~Wang and J.~M. Solomon, ``Deep closest point: Learning representations for point cloud registration,'' in \emph{Proc. {IEEE} Int. Conf. Comput. Vis.}, 2019, pp. 3523--3532.

\bibitem{cao2021pcam}
A.-Q. Cao, G.~Puy, A.~Boulch, and R.~Marlet, ``{PCAM}: Product of cross-attention matrices for rigid registration of point clouds,'' in \emph{Proc. {IEEE} Int. Conf. Comput. Vis.}, 2021, pp. 13\,229--13\,238.

\bibitem{yew2020rpmnet}
Z.~J. Yew and G.~H. Lee, ``Rpm-net: Robust point matching using learned features,'' in \emph{Proc. {IEEE} Conf. Comput. Vis. Pattern Recog.}, 2020, pp. 11\,824--11\,833.

\bibitem{fischler1981ransac}
M.~A. Fischler and R.~C. Bolles, ``Random sample consensus: a paradigm for model fitting with applications to image analysis and automated cartography,'' \emph{Communications of the ACM}, vol.~24, no.~6, pp. 381--395, 1981.

\bibitem{mei2023unsupervised}
G.~Mei, H.~Tang, X.~Huang, W.~Wang, J.~Liu, J.~Zhang, L.~Van~Gool, and Q.~Wu, ``Unsupervised deep probabilistic approach for partial point cloud registration,'' in \emph{Proc. {IEEE} Conf. Comput. Vis. Pattern Recog.}, 2023, pp. 13\,611--13\,620.

\bibitem{aijazi2024non}
A.~K. Aijazi and P.~Checchin, ``Non-repetitive scanning lidar sensor for robust 3d point cloud registration in localization and mapping applications,'' \emph{Sensors}, vol.~24, no.~2, p. 378, 2024.

\bibitem{vaswani2017attention}
A.~Vaswani, N.~Shazeer, N.~Parmar, J.~Uszkoreit, L.~Jones, A.~N. Gomez, {\L}.~Kaiser, and I.~Polosukhin, ``Attention is all you need,'' \emph{Advances in neural information processing systems}, vol.~30, 2017.

\bibitem{zhu2021ropnet}
L.~Zhu, D.~Liu, C.~Lin, R.~Yan, F.~G{\'o}mez-Fern{\'a}ndez, N.~Yang, and Z.~Feng, ``Point cloud registration using representative overlapping points,'' \emph{arXiv preprint arXiv:2107.02583}, 2021.

\bibitem{yew2022regtr}
Z.~J. Yew and G.~H. Lee, ``Regtr: End-to-end point cloud correspondences with transformers,'' in \emph{Proc. {IEEE} Conf. Comput. Vis. Pattern Recog.}, 2022, pp. 6677--6686.

\bibitem{huang2022robust}
X.~Huang, Y.~Wang, S.~Li, G.~Mei, Z.~Xu, Y.~Wang, J.~Zhang, and M.~Bennamoun, ``Robust real-world point cloud registration by inlier detection,'' \emph{Comput. Vis. Image Understanding}, vol. 224, p. 103556, 2022.

\bibitem{yu2023peal}
J.~Yu, L.~Ren, Y.~Zhang, W.~Zhou, L.~Lin, and G.~Dai, ``{PEAL}: Prior-embedded explicit attention learning for low-overlap point cloud registration,'' in \emph{Proc. {IEEE} Conf. Comput. Vis. Pattern Recog.}, 2023, pp. 17\,702--17\,711.

\bibitem{zhao2022g3doa}
H.~Zhao, H.~Zhuang, C.~Wang, and M.~Yang, ``{G3DOA}: Generalizable 3d descriptor with overlap attention for point cloud registration,'' \emph{IEEE Robotics and Automation Letters}, vol.~7, no.~2, pp. 2541--2548, 2022.

\bibitem{deng2018ppfnet}
H.~Deng, T.~Birdal, and S.~Ilic, ``Ppfnet: Global context aware local features for robust 3d point matching,'' in \emph{Proc. {IEEE} Conf. Comput. Vis. Pattern Recog.}, 2018, pp. 195--205.

\bibitem{ao2021spinnet}
S.~Ao, Q.~Hu, B.~Yang, A.~Markham, and Y.~Guo, ``Spinnet: Learning a general surface descriptor for 3d point cloud registration,'' in \emph{Proc. {IEEE} Conf. Comput. Vis. Pattern Recog.}, 2021, pp. 11\,753--11\,762.

\bibitem{yu2023roitr}
H.~Yu, Z.~Qin, J.~Hou, M.~Saleh, D.~Li, B.~Busam, and S.~Ilic, ``Rotation-invariant transformer for point cloud matching,'' in \emph{Proc. {IEEE} Conf. Comput. Vis. Pattern Recog.}, 2023, pp. 5384--5393.

\bibitem{ao2023buffer}
S.~Ao, Q.~Hu, H.~Wang, K.~Xu, and Y.~Guo, ``Buffer: Balancing accuracy, efficiency, and generalizability in point cloud registration,'' in \emph{Proc. {IEEE} Conf. Comput. Vis. Pattern Recog.}, 2023, pp. 1255--1264.

\bibitem{pmlr-v164-zhu22b}
M.~Zhu, M.~Ghaffari, and H.~Peng, ``Correspondence-free point cloud registration with {SO(3)}-equivariant implicit shape representations,'' in \emph{Proceedings of the 5th Conference on Robot Learning}.\hskip 1em plus 0.5em minus 0.4em\relax PMLR, 2022, pp. 1412--1422.

\bibitem{fuchs2020se}
F.~Fuchs, D.~Worrall, V.~Fischer, and M.~Welling, ``{SE}(3)-transformers: 3d roto-translation equivariant attention networks,'' \emph{Proc. Advances Neural Inform. Process. Syst. Conf.}, vol.~33, pp. 1970--1981, 2020.

\bibitem{chatzipantazis2022se}
E.~Chatzipantazis, S.~Pertigkiozoglou, E.~Dobriban, and K.~Daniilidis, ``{SE}(3)-equivariant attention networks for shape reconstruction in function space,'' \emph{arXiv preprint arXiv:2204.02394}, 2022.

\bibitem{wang2023roreg}
H.~Wang, Y.~Liu, Q.~Hu, B.~Wang, J.~Chen, Z.~Dong, Y.~Guo, W.~Wang, and B.~Yang, ``Roreg: Pairwise point cloud registration with oriented descriptors and local rotations,'' \emph{{IEEE} Trans. Pattern Anal. Mach. Intell.}, 2023.

\bibitem{thomas2019kpconv}
H.~Thomas, C.~R. Qi, J.-E. Deschaud, B.~Marcotegui, F.~Goulette, and L.~J. Guibas, ``Kpconv: Flexible and deformable convolution for point clouds,'' in \emph{Proc. {IEEE} Int. Conf. Comput. Vis.}, 2019, pp. 6411--6420.

\bibitem{lin2017feature}
T.-Y. Lin, P.~Doll{\'a}r, R.~Girshick, K.~He, B.~Hariharan, and S.~Belongie, ``Feature pyramid networks for object detection,'' in \emph{Proc. {IEEE} Conf. Comput. Vis. Pattern Recog.}, 2017, pp. 2117--2125.

\bibitem{zhou2018open3d}
Q.-Y. Zhou, J.~Park, and V.~Koltun, ``Open3d: A modern library for 3d data processing,'' \emph{arXiv preprint arXiv:1801.09847}, 2018.

\bibitem{zeng20173dmatch}
A.~Zeng, S.~Song, M.~Nie{\ss}ner, M.~Fisher, J.~Xiao, and T.~Funkhouser, ``3dmatch: Learning local geometric descriptors from {RGB-D} reconstructions,'' in \emph{Proc. {IEEE} Conf. Comput. Vis. Pattern Recog.}, 2017, pp. 1802--1811.

\bibitem{bai2020d3feat}
X.~Bai, Z.~Luo, L.~Zhou, H.~Fu, L.~Quan, and C.-L. Tai, ``D3feat: Joint learning of dense detection and description of 3d local features,'' in \emph{Proc. {IEEE} Conf. Comput. Vis. Pattern Recog.}, 2020, pp. 6359--6367.

\bibitem{choy2019fcgf}
C.~Choy, J.~Park, and V.~Koltun, ``Fully convolutional geometric features,'' in \emph{Proc. {IEEE} Int. Conf. Comput. Vis.}, 2019, pp. 8958--8966.

\bibitem{geiger2012kitti}
A.~Geiger, P.~Lenz, and R.~Urtasun, ``Are we ready for autonomous driving? the kitti vision benchmark suite,'' in \emph{Proc. {IEEE} Conf. Comput. Vis. Pattern Recog.}\hskip 1em plus 0.5em minus 0.4em\relax IEEE, 2012, pp. 3354--3361.

\bibitem{choy2019fully}
C.~Choy, J.~Park, and V.~Koltun, ``Fully convolutional geometric features,'' in \emph{Proc. {IEEE} Int. Conf. Comput. Vis.}, 2019, pp. 8958--8966.

\bibitem{pmlr-v205-lin23a}
C.~E. Lin, J.~Song, R.~Zhang, M.~Zhu, and M.~Ghaffari, ``{SE}(3)-equivariant point cloud-based place recognition,'' in \emph{Proceedings of The 6th Conference on Robot Learning}.\hskip 1em plus 0.5em minus 0.4em\relax PMLR, 2023, pp. 1520--1530.

\end{thebibliography}
}

\end{document}